\documentclass[sigconf]{acmart}
\usepackage{lipsum}
\usepackage{graphicx}
\usepackage{footnote}

\usepackage{amssymb}

\usepackage{footnote}
\usepackage{amsmath}
\usepackage{mathtools}
\usepackage{mathrsfs}                                      
\usepackage{siunitx} 
\usepackage{comment}
\usepackage{wrapfig} 
\usepackage[subrefformat=parens,labelformat=parens]{subfig}
\usepackage{bm,bbm}
\usepackage{multirow}
\usepackage{makecell}
\usepackage{threeparttable,booktabs}
\usepackage{blkarray}
\usepackage{tikz}
\usetikzlibrary{positioning,calc,fit,decorations.pathmorphing,shapes.geometric,shapes.gates.logic.US,calc}
\usepackage{tikz-3dplot}
\usepackage{balance}
\usepackage{courier}                                       
\usepackage{cleveref}                                      
\usepackage[mathcal]{eucal}
\usepackage[]{algpseudocode}                               
\algrenewcommand\textproc{\texttt}
\makeatletter\let\float@addtolists\relax\makeatother
\usepackage{algorithm}
\algblockdefx[pardo]{pardo}{endpardo}%
   [1]{\textbf{For} #1 \textbf{do in parallel}}%
   {\textbf{Endfor}}
\usepackage{filecontents}                                  
\usepackage{pgfplots}
\usepackage{pgfplotstable}
\usepackage{pgf-pie}
\usepgfplotslibrary{groupplots}
\pgfplotsset{compat=newest}
\usepackage[figuresright]{rotating}
\usepackage{xcolor,colortbl}                               
\usepackage[inline]{enumitem}                              
\setlist{leftmargin=5.08mm}

\theoremstyle{plain}

\theoremstyle{definition}

\algrenewcommand\textproc{\texttt}

\usepackage[skip=1pt]{caption}            
\definecolor{CUHKorange}{RGB}{244,106,18} 
\definecolor{CUHKblue}{RGB}{0,111,190}    
\definecolor{CUHKgreen}{RGB}{0,127,128}   
\definecolor{CUHKred}{RGB}{228,46,36}     
\definecolor{CUHKyellow}{RGB}{198,148,34} 
\definecolor{CUHKdark}{RGB}{114,44,114}   
\definecolor{CUHKmiddle}{RGB}{144,44,144} 
\definecolor{red2}{RGB}{192,0,0} 
\usepackage{tcolorbox}
\tcbuselibrary{skins,breakable}
    {\endtcolorbox}

\graphicspath{{./figs/}{./submission/figs/DTCO/}}

\copyrightyear{2024}
\acmYear{2024}
\setcopyright{acmlicensed}\acmConference[ICCAD '24]{IEEE/ACM International Conference on Computer-Aided Design}{October 27--31, 2024}{New York, NY, USA}
\acmBooktitle{IEEE/ACM International Conference on Computer-Aided Design (ICCAD '24), October 27--31, 2024, New York, NY, USA}
\acmDOI{10.1145/3676536.3676750}
\acmISBN{979-8-4007-1077-3/24/10}

\begin{document}
\date{}

\title{
    FabGPT: An Efficient Large Multimodal Model for Complex Wafer Defect Knowledge Queries
}

\author{Yuqi Jiang}
\affiliation{%
  \institution{Zhejiang University}
  \city{Hangzhou}
  \country{China}}

\author{Xudong Lu}
\affiliation{%
  \institution{Zhejiang University}
  \city{Hangzhou}
  \country{China}}

\author{Qian Jin}
\affiliation{%
  \institution{Zhejiang University}
  \city{Hangzhou}
  \country{China}}

\author{Qi Sun$^{\#}$}
\affiliation{%
  \institution{Zhejiang University}
  \city{Hangzhou}
  \country{China}}
\email{qisunchn@zju.edu.cn}

\author{Hanming Wu}
\affiliation{%
  \institution{Zhejiang University}
  \city{Hangzhou}
  \country{China}}

\author{Cheng Zhuo$^{\#}$}
\affiliation{%
  \institution{Zhejiang University}
  \city{Hangzhou}
  \country{China}}
\email{czhuo@zju.edu.cn}

\thanks{$^\#$ Corresponding authors}

\begin{abstract}
Intelligence is key to advancing integrated circuit (IC) fabrication. Recent breakthroughs in Large Multimodal Models (LMMs) have unlocked extraditionary abilities in understanding images and text, fostering intelligent fabrication. 
Leveraging the power of LMMs, we introduce FabGPT, a customized IC fabrication large multimodal model for wafer defect knowledge query. FabGPT manifests expertise in conducting defect detection in Scanning Electron Microscope (SEM) images, performing root cause analysis, and providing expert Q\&A on fabrication processes. FabGPT matches enhanced multimodal features to automatically detect minute defects under complex wafer backgrounds and reduce the subjectivity of manual threshold settings. Besides, the proposed modulation module and interactive corpus training strategy embed wafer defect knowledge into the pre-trained model, effectively balancing Q\&A queries related to defect knowledge and original knowledge and mitigating the modality bias issues. 
Experiments on in-house fab data show that FabGPT achieves significant performance improvement in wafer defect detection and knowledge querying.

\end{abstract}


\maketitle
\pagestyle{plain}

\section{Introduction}
\label{sec:introduction}

The intersection of visual and language models \cite{achiam2023gpt, su2023pandagpt,touvron2023llama} has significantly propelled the revolutionary advancement of artificial intelligence (AI), which makes models understand and interpret the world similarly to humans. Since Large Multimodal Models (LMMs) \cite{zhu2023minigpt, radford2021learning, reddy2021dall} possess the capability to reason about visual images, they have attracted considerable attention in defect detection tasks. However, current LMMs are primarily applied to visual tasks \cite{gu2023pre, chen2022improving, wei2023symbol} in basic scenarios and lack sensitivity to the knowledge of specialized domains. This limits their efficiency in wafer defect knowledge query in the field of integrated circuits (IC) fabrication.

In the semiconductor industry, the manufacturing process is intricate, with each step potentially introducing random defects. These defects impact the reliability of electronic devices \cite{quirk2001semiconductor, fan2019key, lechien2023automated, seebauer2010trends}, making it essential to detect defects on the wafer surface and perform thorough question-and-answer (Q\&A) analysis to deepen engineers' understanding of these defects and IC questions.

\begin{figure}[tb!]
    \centering
    \includegraphics[width=0.96\linewidth]{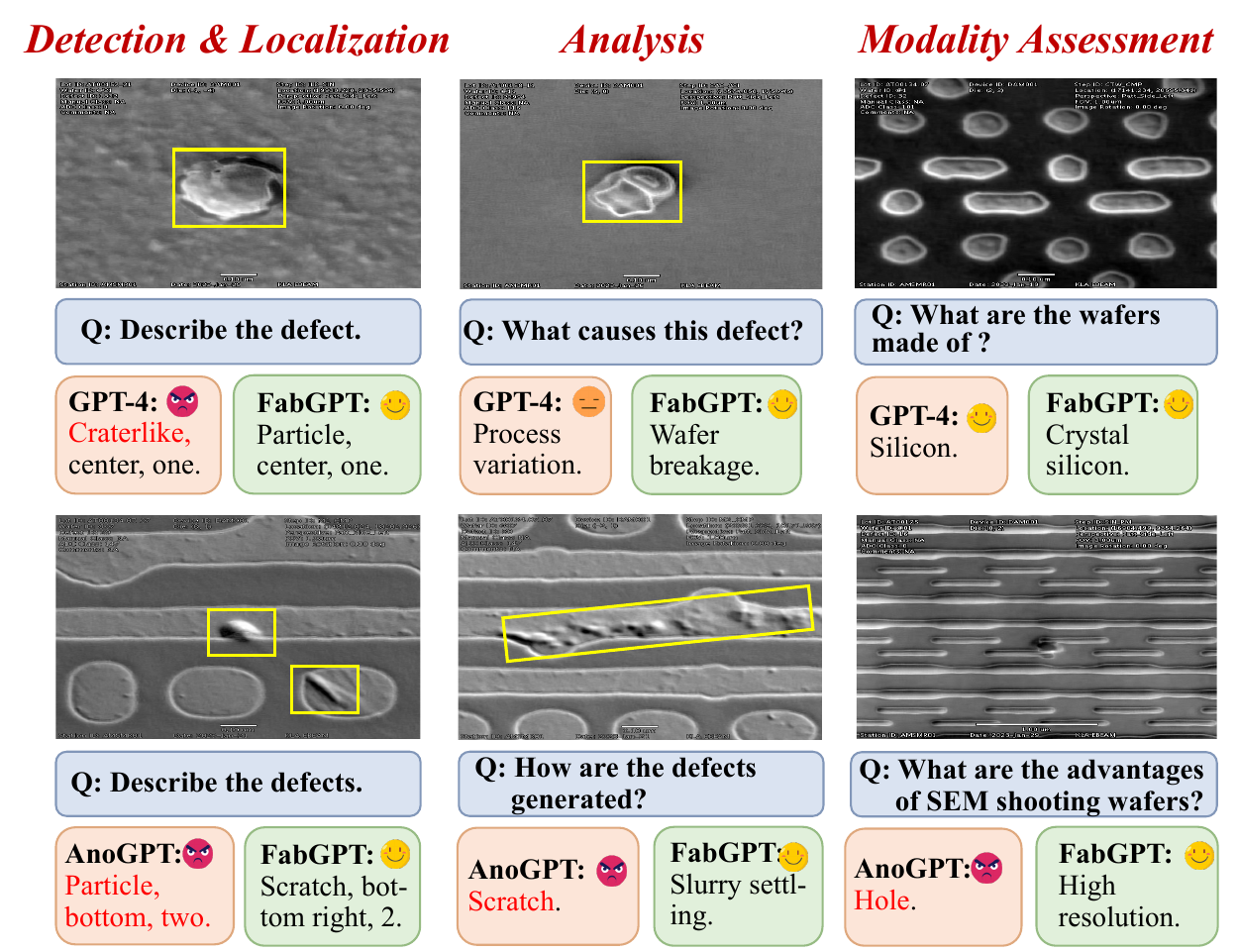}
    \caption{Comparisons between our FabGPT and GPT-4 \cite{achiam2023gpt}, AnomalyGPT \cite{gu2023anomalygpt}, which are fine-tuned on our dataset, for detecting, locating, and analyzing microscopic defects in complex backgrounds and addressing modality bias issues. Previous arts perform badly while encountering ``detection'', ``analysis'', and ``modality bias''. }
    \label{fig:qa-example}
\end{figure}

Recent years have witnessed advancements in methods for querying defect knowledge, encompassing both detection and Q\&A analysis. Convolutional Neural Networks (CNNs)-based approaches \cite{pang2021explainable, cheon2019convolutional, ding2022catching, zhang2023prototypical, yao2023explicit} leverage extensive training data to recognize patterns and features in images, thereby enhancing the accuracy and efficiency of defect detection. However, these methods are heavily dependent on large-scale annotated data and fall short of conducting in-depth Q\&A analysis, which limits their ability to understand complex image content. Moreover, some methods \cite{gu2023anomalygpt, lai2023lisa} employ large models, fine-tuning pre-trained models with specific data sets to achieve superior detection performance, even with limited data availability. These approaches also demonstrate strong visual understanding capabilities, enabling them to deduce relevant visual knowledge. Consequently, integrating wafer defect knowledge from the IC domain into large models is promising to support comprehensive defect knowledge queries.

However, existing methods based on fine-tuning large models often face two issues: 1) difficulty in precisely detecting defects within complex backgrounds; and 2) a loss of the ability to perform comprehensive Q\&A tasks. Despite the vast amount of knowledge stored in large models, the complex and microscopic structure of wafer surfaces still prevents them from accurately capturing information such as the number and location of minute defects. As shown in \Cref{fig:qa-example}, GPT-4 \cite{achiam2023gpt} struggles with queries specific to wafer defects, and even the fine-tuned AnomalyGPT \cite{gu2023anomalygpt} inaccurately identifies the locations of minute defects. Additionally, as demonstrated by the dialogue in \Cref{fig:qa-example} with AnomalyGPT \cite{gu2023anomalygpt}, the fine-tuned model tends to produce text outputs that are still biased towards visual content when user queries are not closely related to the visual input. This is termed as ``modality bias'', indicating that the model loses its ability to judge and understand the textual content of the questions.

To address the issues mentioned above, we propose an efficient LMM, FabGPT, that employs a three-stage strategy: modal enhancement, detection, and Q\&A stage. This strategy allows us to build on the inherent capabilities of pre-trained models by embedding high-quality prompt instructions, enabling it to detect wafer defects in the IC domain and query the related knowledge. Additionally, to further alleviate ``modality bias'', we introduce an interactive corpus training strategy. Our contributions are summarized as follows:
\begin{itemize}
    \item We propose a knowledge query LMM, FabGPT, based on prompt learning, which effectively detects minute defects in complex wafer backgrounds and conducts Q\&A analysis on relevant defect knowledge.
    \item We design a modal enhancement stage that constrains and supplements the semantic information in multimodal features, significantly optimizing the quality of the prompt features.
    \item A detection head is developed, matching multimodal features to automatically detect pixel-level defects. This device eliminates the drawbacks of subjective threshold selection while enhancing defect detection capabilities.
    \item We propose a Q\&A stage and a corpus training strategy that supervises real-time updating of instruction coefficients and the interaction of new and old knowledge, addressing the modality bias in dialogues.
    \item We conduct comprehensive experiments on our SEM-WaD dataset, our FabGPT achieves the supervised detection accuracy of 91.81\% image-level, 95.61\% pixel-level, 88.17\% PRO, and 85.80\% AP. For Q\&A dialogue, it achieves 96.86\% accuracy, outperforming the baselines significantly.
\end{itemize}

\section{Preliminaries}
\label{sec:preliminaries}

\subsection{IC Wafer Defect Analysis}

Wafers are the fundamental material for manufacturing IC, and the quality of their surface directly impacts the reliability of the chips. By identifying and understanding the various defects on the wafer surfaces detected by Scanning Electron Microscopes (SEM) \cite{quirk2001semiconductor,fan2019key,seebauer2010trends}, engineers can learn about the attributes such as type, location, and cause of the defects, which is crucial for optimizing processes and quality control. Traditional and CNN-based analysis methods utilize large datasets for feature learning to detect defect regions. For example, Zontak \textit{et al.} \cite{zontak2009kernel} utilized the periodicity of wafer patterns to manually construct defect features, thus detecting defects. Gomez \textit{et al.} \cite{gomez2022optimal} proposed a detection method based on Support Vector Machines (SVM), which separates data points of different classes in high-dimensional space using defined hyperplanes. Cheon \textit{et al.} \cite{cheon2019convolutional} proposed a CNN model that can extract effective features for defect classification. These models achieve significant progress in classifying and segmenting wafer defects. However, they perform poorly when faced with scarce data and have not yet developed a model that integrates classification, segmentation, and analysis. There is an urgent need for an automated tool capable of detecting and inferring wafer defects with minimal data.

\begin{figure}[tb!]
    \centering
    \includegraphics[width=0.96\linewidth]{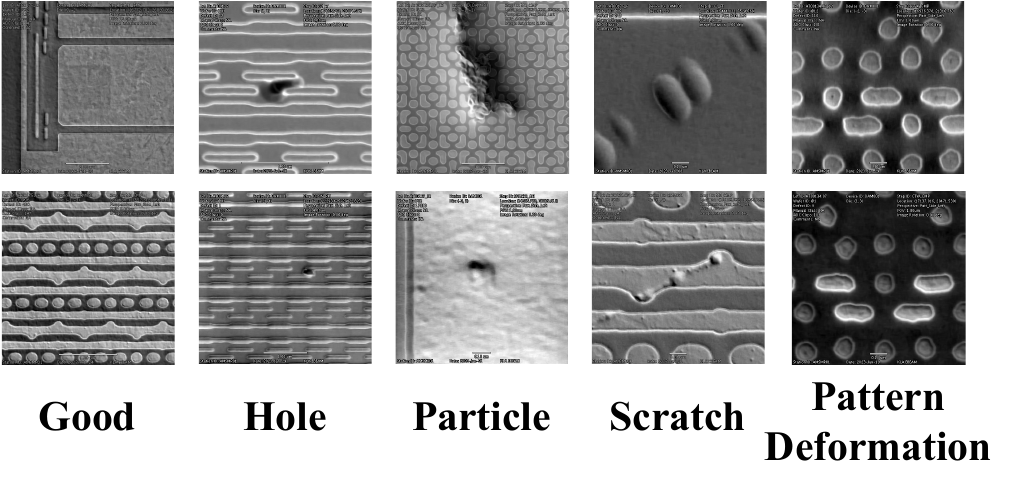}
    \caption{The four types of defects and defect-free (good) images in the SEM-WaD dataset.}
    \label{fig:2}
\end{figure}

We use an in-house dataset, SEM-WaD, which comprises 1,226 defect-free images and 1,182 images with four common types of defects (holes, particles, pattern deformities, and scratches). The wafer images contain diverse and complex backgrounds, with each defect type exhibiting unique morphologies and features.  Some examples are shown in \Cref{fig:2}. 

\begin{figure*}[ht]
    \centering
    \includegraphics[width=0.88\textwidth]{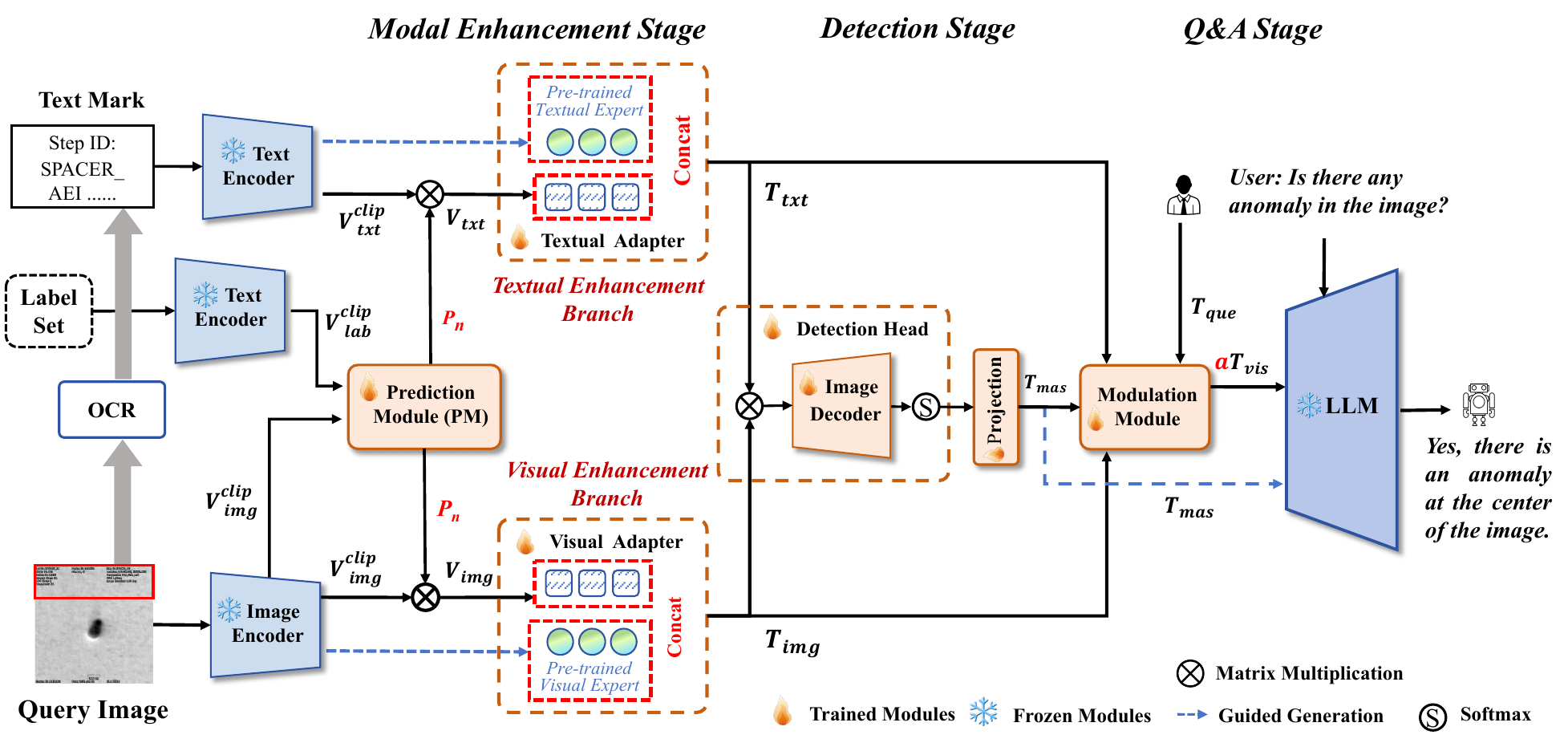}
    \caption{The architecture of FabGPT. The images and the characters extracted from them serve as the primary multimodal input into a three-stage model, with the label set entering as auxiliary textual input. The first stage enhances semantic information of multimodal features. Based on the first stage, the detection stage performs pixel-level automated detection, and the Q\&A stage achieves complete Q\&A integrating both old and new knowledge.}
    \label{fig: network}
\end{figure*}

\subsection{Large Multimodal Models}

LMMs \cite{zhu2023minigpt, radford2021learning, reddy2021dall} are the large and complex models capable of processing various types of data (images, spectra, sound, \textit{etc.}) and utilize large-scale datasets during training to understand and generate descriptions of visual content. Their powerful comprehension and transfer abilities make them excel in various tasks such as image description generation and visual question answering. For example, \cite{zhu2023minigpt} utilized a linear layer for aligning the frozen video encoder of the BLIP-2 \cite{li2023blip} and the LLM Vicuna \cite{chiang2023vicuna} to enable image-text Q\&A. \cite{radford2021learning} adopted a contrastive learning approach to embed the semantic information of images and text into the same space to achieve zero-shot transfer. DALL-E \cite{reddy2021dall} generated images related to text descriptions by encoding text with Transformer and using a Generative Adversarial Network (GAN). These models have a strong ability to understand complex image-text data, however, they face challenges in robustness when adapting to new domains.

\subsection{Fine-Tuning Methods}

Fine-tuning methods primarily include full \cite{alaparthi2020bidirectional, yuan2021tokens} and partial fine-tuning \cite{rebuffi2017learning, lai2023lisa, gu2023anomalygpt}. In full fine-tuning, all pre-trained model layer weights are trainable, enabling flexible adaptation to new data features by adjusting all parameters. For example, BERT \cite{alaparthi2020bidirectional} is trained on large corpora adjusting all parameters in its bidirectional Transformer structure to optimize tasks such as sentiment analysis question answering and text summarization. 
Full fine-tuning improves task adaptation but increases overfitting risk and costs, limiting its use. 

In partial fine-tuning \cite{rebuffi2017learning, lai2023lisa, gu2023anomalygpt}, it adjusts only partial parameters in pre-trained models and keeps the rest fixed. Recent popular partial fine-tuning strategies include Adapters, Prompt Learning, and LoRA. For example, Lai \textit{et al.} \cite{lai2023lisa} applied the LoRA method to consistently update the bottom embeddings and top linear head of the pre-trained model, while randomly updating a few intermediate self-attention layers to understand input text for precise segmentation. Gu \textit{et al.} proposed AnomalyGPT \cite{gu2023anomalygpt}, which fine-tunes an LLM using embedded prompt instructions to identify types and locations of defects. 

Despite their advancements, new models often lose their normal Q\&A capabilities after fine-tuning. This occurs because the model becomes excessively focused on image inputs while neglecting user queries, whether or not these queries are related to the images. This phenomenon is termed ``\textit{\textbf{modality bias}}''. Our model effectively integrates complex wafer defect knowledge and is designed to alleviate the modality bias.

\section{Proposed Method}
\label{sec: method}

\subsection{Network Architecture}

As shown in \Cref{fig: network}, our FabGPT is a conversational LMM designed for querying wafer defect knowledge, and it consists of a foundational stage for modal enhancement and two functional stages for detection and Q\&A.

Given a query image $x \in \mathbb{R}^{H \times W \times 3}$, text marks are extracted from $x$ using Optical Character Recognition (OCR) technology \cite{li2022pp}. The image $x$, its text marks, and the label set are encoded into initial vectors $V_{img}^{clip}$, $V_{txt}^{clip}$, and $V_{lab}^{clip}$ through pre-trained image and text encoders \cite{girdhar2023imagebind}. $V_{img}^{clip}$ and $V_{lab}^{clip}$ are fed into the Prediction Module (PM) to predict the defect category $P_n$. Then, $P_n$ is used to multiply with $V_{img}^{clip}$ and $V_{txt}^{clip}$, generating the vectors $V_{img}$ and $V_{txt}$. Visual and textual adapters further process these vectors into the information-rich image token $T_{img}$ and text token $T_{txt}$.
In the detection stage, $T_{img}$ and $T_{txt}$ are fed into the detection head to obtain supervised detection masks. In the Q\&A stage, the Modulation Module aligns $T_{img}$, $T_{txt}$, mask-projected token $T_{mas}$, and the user's question token $T_{que}$ into a unified visual token $aT_{vis}$. Finally, it is concatenated with $T_{mas}$ and $T_{que}$ to serve as prompt instructions for fine-tuning PandaGPT \cite{su2023pandagpt}.

\subsection{Modal Enhancement Stage}

When analyzing the minor defects on the wafer surface, distinguishing between complex background and foreground features is challenging, which hinders queries on defect-related knowledge. To address this, we develop the modal enhancement stage, consisting of the PM and two enhancement branches, designed to highlight relevant defect features and minimize the impact of irrelevant features.

\begin{figure}[htbp]
    \centering
    \includegraphics[width=\linewidth]{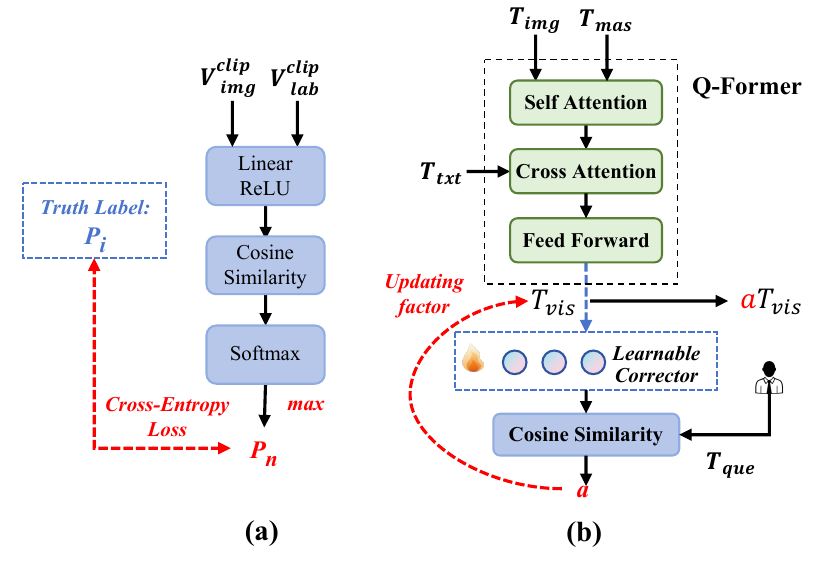}
    \caption{(a) The Prediction Module (PM), (b) The Modulation Module.}
    \label{fig:PM-Modulation}
\end{figure}

\noindent\textbf{Prediction Module (PM)}:

The pre-trained encoder captures pixel-level detail features in the latent space, but its repeated down-sampling operations result in the loss of semantic features. As shown in \Cref{fig:PM-Modulation} (a), we design the PM to predict defect categories in the image and enhance semantic features by embedding the expected result $p_n$ into the initial vectors. 

Specifically, we count a label set containing all defect categories found in wafer images to support automated classification tasks. First, the linear layer and the activation function are applied to reshape the dimensions of $V_{img}^{clip}$ and $V_{lab}^{clip}$, formulated as:
\begin{equation}
\begin{aligned}
f_{img}&= \sigma(W_i^T V_{img}^{clip} + b_i),\\
f_{lab}&= \sigma(W_i^T V_{lab}^{clip} + b_i),
\end{aligned}
\end{equation}
where $W_i$ represents the weight matrix, $b_i$ represents the bias vector, and $\sigma$ represents the ReLU function. Then, the cosine of the angle between each category $f_{lab}^i$ and $f_{img}$ is computed to assess their similarity, and the cross-entropy loss function is used to constrain the selection of the corresponding category $p_n$.
This process is formulated as:
\begin{equation}
\begin{aligned}
p_i &= \text{Cosine}(f_{img}, \ f_{lab}), \\
P_n &= max(Softmax(p_i) ) ,\\
    &= max( \frac{exp({p_i})}{\sum_{j=1}^{N} exp({p_j})}  ),
\end{aligned}
\end{equation}
where $\text{Cosine}$ represents the cosine similarity calculation. $p_n$ is matrix-multiplied with $V_{img}^{clip}$ and $V_{lab}^{clip}$, resulting in vectors $V_{img}$ and $V_{txt}$, which are enriched with semantic features.

\noindent\textbf{Two Enhancement Branches}:

Although semantic features related to defect attributes are enhanced, the detailed representation of defect features remains essential. In the visual and textual enhancement branches, adapters based on prompt learning are deployed, utilizing the extra prompts of pre-trained experts for adaptive feature optimization. The experts are initialized under the guidance of $V_{img}^{clip}$ and $V_{txt}^{clip}$, which enable them to acquire knowledge from image and text modals. During training, they adaptively update parameters and interact with $V_{img}$ and $V_{txt}$ after each update, effectively controlling the direction and quality of the feature flow.
The final outputs $T_{img}$ and $T_{txt}$ are generated, with $T_{img}$ being represented by the following formula (the same applies to $T_{txt}$):
\begin{equation}
\begin{aligned}
\nabla f_{e} &= (V_{img}^{clip})^i \rightarrow z^{i} , \\
T_{img} &= Concat( \nabla f_{e} , \ V_{img}),
\end{aligned}
\end{equation}
where $\rightarrow$ represents the feature-guiding operation, $z^{i}$ and $(V_{img}^{clip})^i$ are the $i$-th elements of the random vector $z$ and $V_{img}^{clip}$ respectively, and $\nabla f_{e}$ is the prompt feature of the pre-trained expert.


\subsection{Detection Stage}
Manually setting segmentation thresholds is not necessarily optimal while continuous adjustments should be made for specific tasks.
We design the detection head that autonomously learns the specific thresholds for each pixel at feature positions to generate pixel-level masks $\text{Mask}\in \mathbb{R}^{H\times W}$. It first fuses complementary information from $T_{img}$ and $T_{txt}$ through matrix multiplication, then maps them back to high-dimensional features through four up-sampling operations of the trainable decoder, and normalizes the output into the mask image through the softmax function. The detection head matches multimodal information and supervises detail features, enabling precise defect detection in complex wafer backgrounds. This process is formulated as:
\begin{equation}
\begin{aligned}
\text{Mask} = & Softmax(( T_{img} \otimes T_{txt} ) \uparrow_s^4),
\end{aligned}
\end{equation}
where $(\cdot)\uparrow_s^4$ represents performing four times up-sampling.

\subsection{Q\&A Stage}

In fine-tuning the large models based on prompt instructions, the commonly used embedded instruction format is:
\begin{equation}
\begin{aligned}
\text{INS} = Concat( x ,\ T_{\text{img}} ,\ T_{\text{que}}).
\label{eq: QA1}
\end{aligned}
\end{equation}

However, this embedding format may lead to modality bias, where the model is dominated by image inputs and fails to respond to questions appropriately. For example, the latest defect Q\&A model AnomalyGPT \cite{gu2023anomalygpt} embeds industrial defect knowledge into its pre-trained model based on \Cref{eq: QA1}, it can only answer questions related to defects, such as:
\begin{itemize}
    \item ``Is there a defect in the image?''
    \item ``Where is the defect located in the image?''
\end{itemize}
However, it fails to answer general questions not closely related to the images, such as:
\begin{itemize}
    \item ``What impact does this defect have on the production line?''
    \item ``What are the core process steps in IC manufacturing?''
\end{itemize}

This phenomenon indicates that while the model gains new information, it loses the understanding of its original knowledge, thereby diminishing its ability to analyze general knowledge effectively.

We suggest that this phenomenon occurs because the model fails to judge the correlation between the query image and the user's question adequately. The keys to resolving this issue and ensuring accurate model responses are: 1) Improving the quality of visual instructions; 2) Enhancing the ability to assess the relevance between visual prompt instructions and user query ones; 3) Optimizing the training strategies of the corpus.

\noindent\textbf{Modulation Module}:

The quality of prompts impacts the model's understanding and responses, so aligning multimodal features is needed to capture comprehensive information and reduce the fine-tuning burden. Inspired by Q-Former \cite{li2023blip}, a bidirectional self-attention in \Cref{fig:PM-Modulation} (b) allows $T_{img}$ to absorb semantic and detailed information from $T_{mas}$ to obtain $f_{i \sim m}$, facilitating interaction within the same modality, formulated as:
\begin{equation}
\begin{aligned}
\text{M}_{img} &= Softmax ( ((T_{img} \ast k_1) (T_{img} \ast k_2) ^T) / \sqrt{d_k} ), \\
\text{M}_{mak} &= Softmax ( ((T_{mas} \ast k_1) (T_{mas} \ast k_2) ^T) / \sqrt{d_k} ), \\
f_{i \sim m} &= \frac{(\text{S}_{i} + \text{S}_{m})}{2} \ T_{img} ,
\end{aligned}
\end{equation}
where $\ast$ represents convolution operations, $k_i$ represents different kernels, and $d_k$ represents the dimension of the feature vector.
Next, aligning the fine-grained information between visual features $f_{i \sim m}$ and textual tokens $T_{txt}$ through cross-attention \cite{huang2019ccnet} allows for the sharing of complementary knowledge across multimodalities, the result $f_{i \sim m \sim t}$ can be formulated as:
\begin{equation}
\begin{aligned}
\text{M}_{i \sim m \sim t} &= Softmax (((f_{i \sim m} \ast k_1) (T_{txt} \ast k_2) ^T)/ \sqrt{d_k}), \\
f_{i \sim m \sim t}  &= \text{M}_{i \sim m \sim t} \ T_{txt}.
\end{aligned}
\end{equation}
Finally, to maintain semantic consistency between the LLM and outputs of the modal enhancement stage, the feed-forward network maps the unified features $f_{i \sim m \sim t}$ to a high-dimensional space and outputs high-quality prompt instructions $T_{vis}$ through the activation of nonlinear layers.

Since the content of user queries involves knowledge of different tasks, we must assess the relationship between query and visual instructions before fine-tuning the LLM. We set a scaling factor $a$, dynamically adjusting its value through learning the association between instructions (the higher the value, the stronger the association). A learnable corrector that is generated under the guidance of $T_{vis}$ is introduced, and its similarity score with the query is calculated to simulate the value of $a$. It can be formulated as:
\begin{equation}
\begin{aligned}
a & = \frac{\nabla f_c\cdot T_{que}}{||\nabla f_c||\cdot ||T_{que}||} ,
\end{aligned}
\end{equation}
where $\cdot$ represents the dot product and $\nabla f_c$ represents the prompt features of the learnable corrector. We assign $a$ as the coefficient to $T_{vis}$. The updated $aT_{vis}$ along with $T_{mas}$ and $T_{que}$ serve as our input instructions, formatted as follows:
\begin{equation}
\begin{aligned}
\widetilde{\text{INS}} &= Concat(a T_{vis}, \ T_{mas}, \ T_{que}) .
\end{aligned}
\end{equation}

\noindent\textbf{Corpus Training Strategy}:

During the corpus training process, the alternating training strategy balances learning new and old knowledge. We establish two corpora: Corpus-A, which includes 15 Q\&A pairs for each category related to defect type, quantity, location, description, and analysis (e.g., Q: What type of defect is in the image? A: The defect in the image is {object}.), and Corpus-B, which contains 100 Q\&A pairs unrelated to defect knowledge (e.g., Q: What is the capital of China? A: The capital of China is Beijing.). Our model trains alternately on these corpora at a 2:1 ratio to prevent it from favoring the retrieval of new knowledge when understanding questions.

\subsection{Loss Functions}

We employ three loss functions to constrain detection and dialogue processes. Focal Loss \cite{lin2017focal} and Dice Loss \cite{milletari2016v} are used to improve the model's segmentation and localization abilities, and Cross-Entropy Loss is used to improve the model's classification and Q\&A ones.

\noindent\textbf{Focal Loss}:

Focal Loss \cite{lin2017focal} aims to address the issue of class imbalance. It introduces a modulation factor $\gamma$ to reduce the relative loss of correctly classified pixels and focus on hard-to-classify and misclassified pixels. It is realized as \Cref{eq: focal}:
\begin{equation}
\begin{aligned} \label{eq: focal}
L_{\text{focal}} = -\frac{1}{ \text{H}\times \text{W}} \sum_{i=1}^{\text{H}\times \text{W}} (1 - p_i)^\gamma \log(p_i),
\end{aligned}
\end{equation}
where $p_i$ represents the probability of the pixel belonging to the correct category, and based on  \cite{gu2023anomalygpt}, we set the $\gamma$ to 2. 

\noindent\textbf{Dice Loss}:

Dice Loss \cite{milletari2016v} aims to maximize the overlap between outputs and actual labels, encouraging the model to learn to produce results closer to the ideal segmentation. It is realized as \Cref{eq: dice}:
\begin{equation}
\begin{aligned} \label{eq: dice}
L_{\text{dice}} = - \frac{\sum_{i=1}^{\text{H}\times \text{W}} y_i \hat{y}_i}{\sum_{i=1}^{\text{H}\times \text{W}} y_i^2 + \sum_{i=1}^{\text{H}\times \text{W}} \hat{y}_i^2},
\end{aligned}
\end{equation}
where $y_i$ is the output of the decoder, and $\hat{y}_i$ is the truth labels.

\noindent\textbf{Cross-Entropy Loss}:

Cross-entropy loss measures the difference between predicted and actual categories in the PM and between output texts of the language model and target texts in the Q\&A task. It is realized as \Cref{eq: cross}:
\begin{equation}
\begin{aligned} \label{eq: cross}
L_{\text{ce}} = - \sum_{i=1}^{c} y_i \log(p_i),
\end{aligned}
\end{equation}
where $c$ represents the number of categories and tokens in classification and Q\&A tasks, $y_i$ represents the truth label and $p_i$ represents the predicted label.

The overall loss function is:
\begin{equation}
\begin{aligned}
L = \alpha L_{focal} + \beta L_{dice} + \delta L_{ce}^1 + \epsilon L_{ce}^2.
\end{aligned}
\end{equation}
We set the coefficients $\alpha$, $\beta$, $\delta$ and $\epsilon$ to 1, by default.

\section{Experiments}
\label{sec:experiments}

\subsection{Experimental Setups}

\begin{table*}[tb!]
  \centering
  \caption{The fully-supervised experimental results. The best and second-best methods are highlighted in \textcolor{red}{red} and \textcolor{blue}{blue}.}
    \resizebox{\linewidth}{!}
    {
        \begin{threeparttable}
        {
            \begin{tabular}{c|ccccc|ccccc}
            \toprule
            \multirow{3}{*}{Methods} & \multicolumn{5}{c|}{Image-AUC / Pixel-AUC} & \multicolumn{5}{c}{PRO / AP} \\
            \cmidrule(lr){2-6} \cmidrule(lr){7-11}
            & \multirow{2}{*}{Hole} & \multirow{2}{*}{Particle} & \multirow{2}{*}{Scratch} & Pattern & \multirow{2}{*}{Average} & \multirow{2}{*}{Hole} & \multirow{2}{*}{Particle} & \multirow{2}{*}{Scratch} & Pattern & \multirow{2}{*}{Average} \\
            & & & & Deformation & & & & & Deformation & \\
            \midrule
            DevNet \cite{pang2021explainable} & 63.92 / 80.10 & 62.20 / 86.30 & 19.84 / 76.74 & 65.02 / 48.78 & 52.74 / 72.98 & \ \ \ \ - \ \ \ / \textcolor{red}{86.98} & \ \ \ \ - \ \ \ / \textcolor{red}{92.38} & \ \ \ \ - \ \ \ / 50.55 & \ \ \ \ - \ \ \ / \textcolor{blue}{82.07} & \ \ \ \ - \ \ \ / \textcolor{blue}{77.80} \\
            DRA  \cite{ding2022catching} & 81.44 / 85.75 & 96.71 / \textcolor{blue}{96.86} & \textcolor{blue}{89.94} / 75.28 & 85.52 / \textcolor{blue}{90.71} & 88.14 / 87.15 & 57.81 / \ \ \ \ - \ \ \ & 66.67 / \ \ \ \ - \ \ \ & 25.83 / \ \ \ \ - \ \ \ & 60.71 / \ \ \ \ - \ \ \ & 52.76 / \ \ \ \ - \ \ \ \\
            BGAD \cite{yao2023explicit} & 35.78 / 87.10 & 81.87 / 92.75 & 65.27 / 88.98 & 72.37 / 81.90 & 63.82 / 87.68 & 80.81 / 51.33 & 89.58 / 51.33 & 75.33 / 52.28 & 81.08 / 54.75 & 81.70 / 52.66 \\
            PRN  \cite{zhang2023prototypical} & 79.26 / 84.30 & 80.41 / 87.60 & 76.93 / 77.02 & 75.87 / 76.53 & 78.12 / 81.36 & 80.90 / 76.48 & 88.84 / 84.79 & 67.82 / 51.70 & 74.63 / 70.57 & 78.05 / 70.89 \\
            Lisa  \cite{lai2023lisa} & 87.37 / 84.31 & \textcolor{blue}{93.34} / 91.73 & 84.45 / 85.28 & 85.51 / 87.72 & 87.67 / 87.26 & 88.66 / 79.18 & 83.57 / 72.87 & 57.77 / 35.73 & 77.75 / 69.01 & 76.94 / 64.20 \\
            AnomalyGPT  \cite{gu2023anomalygpt} & \textcolor{blue}{91.37} / \textcolor{blue}{93.68} & 92.64 / 94.39 & 89.58 / \textcolor{blue}{92.58} & \textcolor{blue}{86.23} / 89.66 & \textcolor{blue}{89.96} / \textcolor{blue}{92.58} & 86.39 / 83.47 & 89.74 / 86.02 & \textcolor{blue}{80.32} / \textcolor{blue}{60.32} & \textcolor{blue}{84.35} / 80.73 & \textcolor{blue}{85.20} / 77.64 \\
            \midrule
            \textbf{FabGPT (ours)} & \textcolor{red}{94.28} / \textcolor{red}{97.03} & \textcolor{red}{94.43} / \textcolor{red}{97.30} & \textcolor{red}{90.32} / \textcolor{red}{95.70} & \textcolor{red}{88.19} / \textcolor{red}{92.40} & \textcolor{red}{91.81} / \textcolor{red}{95.61} & \textcolor{red}{90.01} / \textcolor{blue}{85.69} & \textcolor{red}{92.28} / \textcolor{blue}{91.58} & \textcolor{red}{83.00} / \textcolor{red}{79.19} & \textcolor{red}{87.40} / \textcolor{red}{86.72} & \textcolor{red}{88.17} / \textcolor{red}{85.80} \\
            \bottomrule
            \end{tabular}
        }
        \end{threeparttable}
    }
  \label{tab:merged_metrics}
\end{table*}

\noindent\textbf{Datasets}:

Our experiment is conducted on the in-house SEM-WaD dataset. The dataset comprises images with a resolution of $480\times480$, each accompanied by a corresponding mask image and related textual descriptions and analyses. We divide the data into training and test sets in a 7:3 ratio, both containing good and defective images. 

\noindent\textbf{Implementation Details}:

Our model uses PandaGPT \cite{su2023pandagpt} as the foundational LMM, composed of the Vicuna-7B \cite{chiang2023vicuna} as the language model and ImageBind Huge \cite{girdhar2023imagebind} as the frozen encoder. During training, we employ the AdamW optimizer ($\beta_1$ = 0.9, $\beta_2$ = 0.999) \cite{loshchilov2017decoupled} with an initial learning rate of $1e^{-4}$, gradually reducing to $1e^{-6}$, using the cosine annealing \cite{loshchilov2016sgdr} strategy. The model is trained on three 4090Ti GPUs, with a batch size of 24 and an epoch of 50.

\noindent\textbf{Evaluations}:

In the detection task, we evaluate model performance using four metrics: Image-AUC (the Area Under the Receiver Operating Characteristic Curve), Pixel-AUC, Per-Region Overlap (PRO), and Average Precision (AP). Image-AUC and Pixel-AUC assess the model's ability to judge the presence of defects in images, while PRO and AP measure the precision of the model in identifying and locating defects. In the Q\&A task, we conduct 15 questions for each of the four defect types, including inquiries about the presence, category, location, quantity, appearance description, and root cause analysis of defects. Additionally, we pose 50 questions that are unrelated to defects and not included in the corpus (IC-related or IC-unrelated general questions) to validate the modality bias issue. We use the percentage of correct answers as a relevant metric to evaluate the Q\&A capability of the model. To demonstrate the outstanding performance of our FabGPT, we compare it with many representative previous arts:
\begin{itemize}
    \item DevNet \cite{pang2021explainable}: Learns anomaly representations by labeled anomalies and prior probabilities.
    \item DRA \cite{ding2022catching}: A CNN-based learning model for detecting anomalies in a composite feature space.
    \item BGAD \cite{yao2023explicit}: An anomaly scoring model that utilizes explicit boundary generation and boundary-guided optimization.
    \item PRN \cite{zhang2023prototypical}: A residual detection model that outputs anomalies by learning different block features. 
    \item Lisa \cite{lai2023lisa}: An anomaly segmentation LMM utilizing LoRA fine-tuning method.
    \item AnomalyGPT \cite{gu2023anomalygpt}: An anomaly detection LMM utilizing prompting learning fine-tuning method.
\end{itemize} 

These baselines are also tuned on our SEM-WaD dataset, following their publicly available implementations and models. Among these baselines, Lisa \cite{lai2023lisa} and AnomalyGPT \cite{gu2023anomalygpt} are LMMs that support Q\&A while the others do not.  

\subsection{Quantitative Results}


\noindent\textbf{Defect Detection Task}:

\Cref{tab:merged_metrics} report the Image-AUC, Pixel-AUC, PRO, and AP values for different methods across 4 categories within the SEM-WaD dataset. It can be observed that our model outperforms all other methods in four evaluation metrics for most defect categories. For example, compared to AnomalyGPT \cite{gu2023anomalygpt}, which also employs prompt learning for fine-tuning, our model achieves a higher Image-AUC by 1.85\% and a higher Pixel-AUC by 3.03\%. Compared to the traditional detection model PRN \cite{zhang2023prototypical}, our model surpasses it by 10.12\% in PRO and 14.91\% in AP.

\begin{table}[h]
\caption{The accuracies (\%) of the models' responses to various questions. ``-'' denotes the question is unsupported. }
\centering
\resizebox{0.88\linewidth}{!}
{
\begin{threeparttable}
{
\begin{tabular}{c|cccc}
\midrule
Questions & Lisa \cite{lai2023lisa} & AnomalyGPT \cite{gu2023anomalygpt} & GPT-4 \cite{achiam2023gpt} & \textbf{FabGPT} \\ \midrule
Presence      & 95.00 & 95.00 & \textcolor{red}{100.00} & \textcolor{red}{100.00} \\
Category      & \textcolor{blue}{92.50} & 75.00 & 37.50 & \textcolor{red}{97.50} \\
Location      & - & 72.50 & \textcolor{blue}{87.50} & \textcolor{red}{95.00} \\
Quantity      & - & 77.50 & \textcolor{blue}{80.00} & \textcolor{red}{95.00} \\
Description   & 72.50 & 82.50 & \textcolor{blue}{85.00} & \textcolor{red}{95.00} \\
Analysis      & - & - & \textcolor{blue}{90.00} & \textcolor{red}{97.50} \\
Unrelated     & 20.00 & 12.00 & \textcolor{red}{98.00} & \textcolor{red}{98.00} \\ \midrule
All           & 70.00 & 69.08 & \textcolor{blue}{82.57} & \textcolor{red}{96.86} \\ 
\midrule
\end{tabular}
}
\end{threeparttable}
}
\label{tab:QA}
\end{table}

\begin{figure*}[tb!]
    \centering
    \begin{minipage}[b]{0.56\textwidth}
        \centering
        \includegraphics[height=0.6\linewidth]{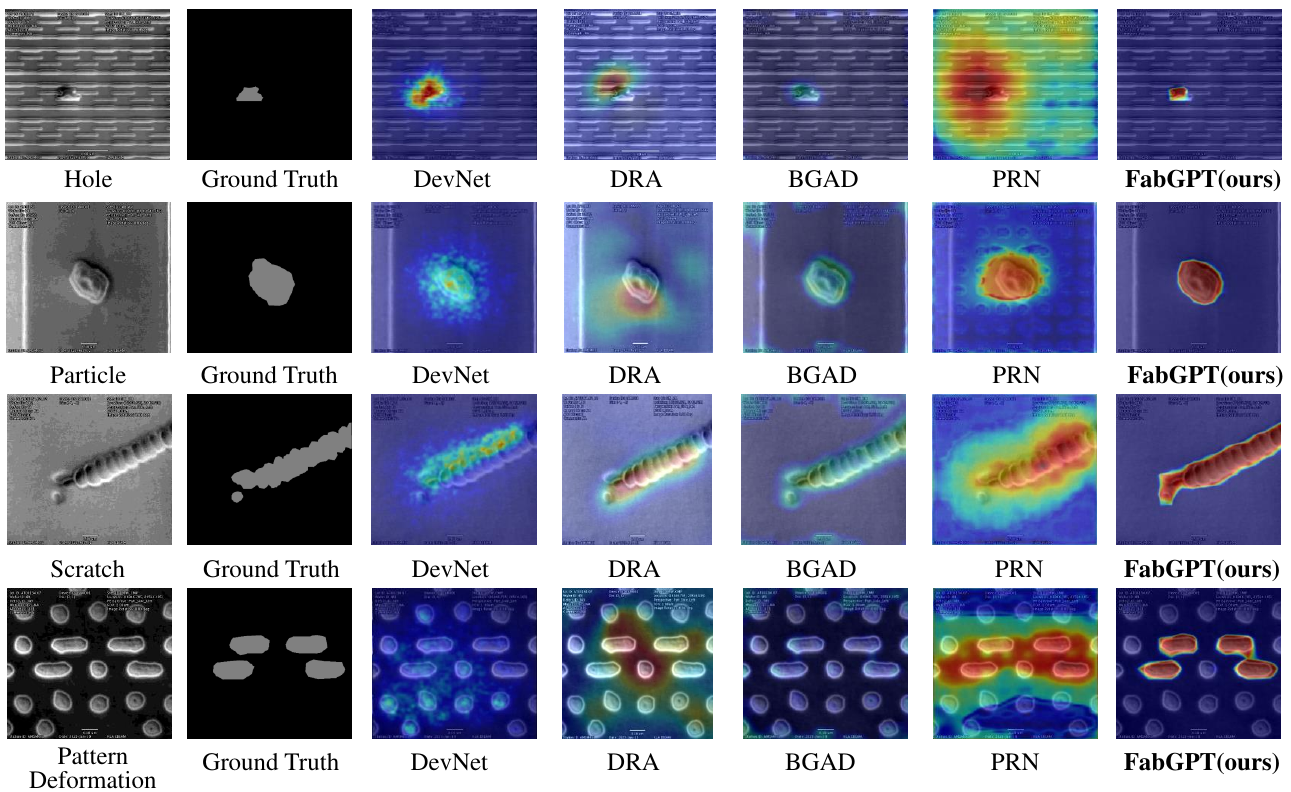}
        \caption{Comparisons with non-LMM baselines.}
        \label{fig:hotmap}
    \end{minipage}
    \hfill
    \begin{minipage}[b]{0.42\textwidth}
        \centering
        \includegraphics[height=0.81\linewidth]{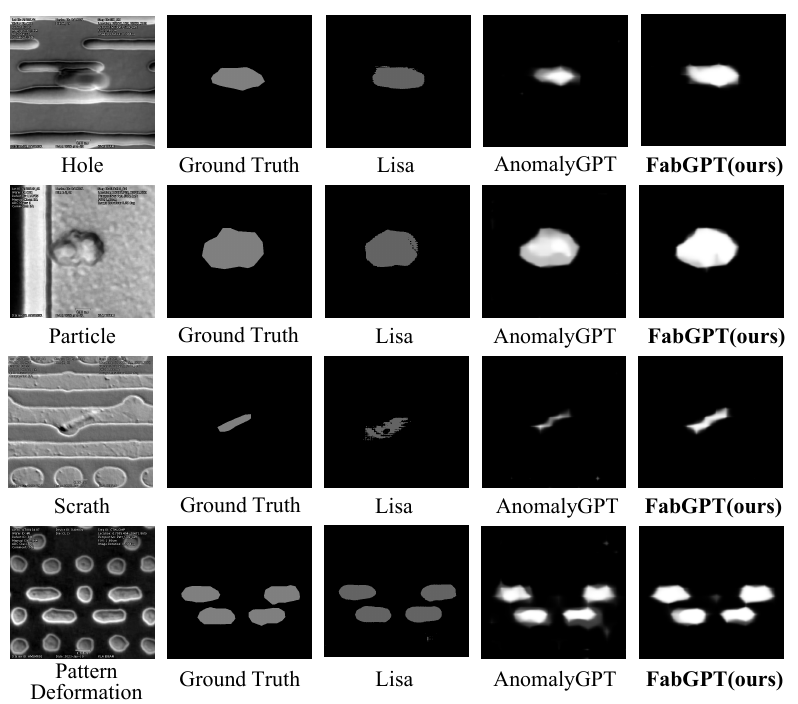}
        \caption{Comparisons with LMM-based baselines.}
        \label{fig:mask}
    \end{minipage}
\end{figure*}

\noindent\textbf{Q\&A Task}:

\Cref{tab:QA} reports the accuracy of different language models in answering defect-related and -unrelated questions where our model achieves state-of-the-art results. For example, compared to the powerful capabilities of GPT-4, our model achieves a comparable performance in the general questions and a 7.50\% higher accuracy in defect analysis answers. Compared to Lisa \cite{lai2023lisa}, which embeds new tasks to pre-trained models, our model achieves a 78.00\% higher accuracy in answering questions unrelated to the new task. 

\subsection{Qualitative Results}

\noindent\textbf{Defect Detection Task}:

Comparing heatmaps in \Cref{fig:hotmap}, we observe that our model focuses more on defect regions than others, while avoiding concentrating too much on normal regions. 
Moreover, comparing mask images in \Cref{fig:mask}, FabGPT provides more accurate edge segmentation details compared to both Lisa \cite{lai2023lisa} and AnomalyGPT \cite{gu2023anomalygpt}. 

\begin{figure}[tb!]
    \centering
    \includegraphics[width=0.88\linewidth]{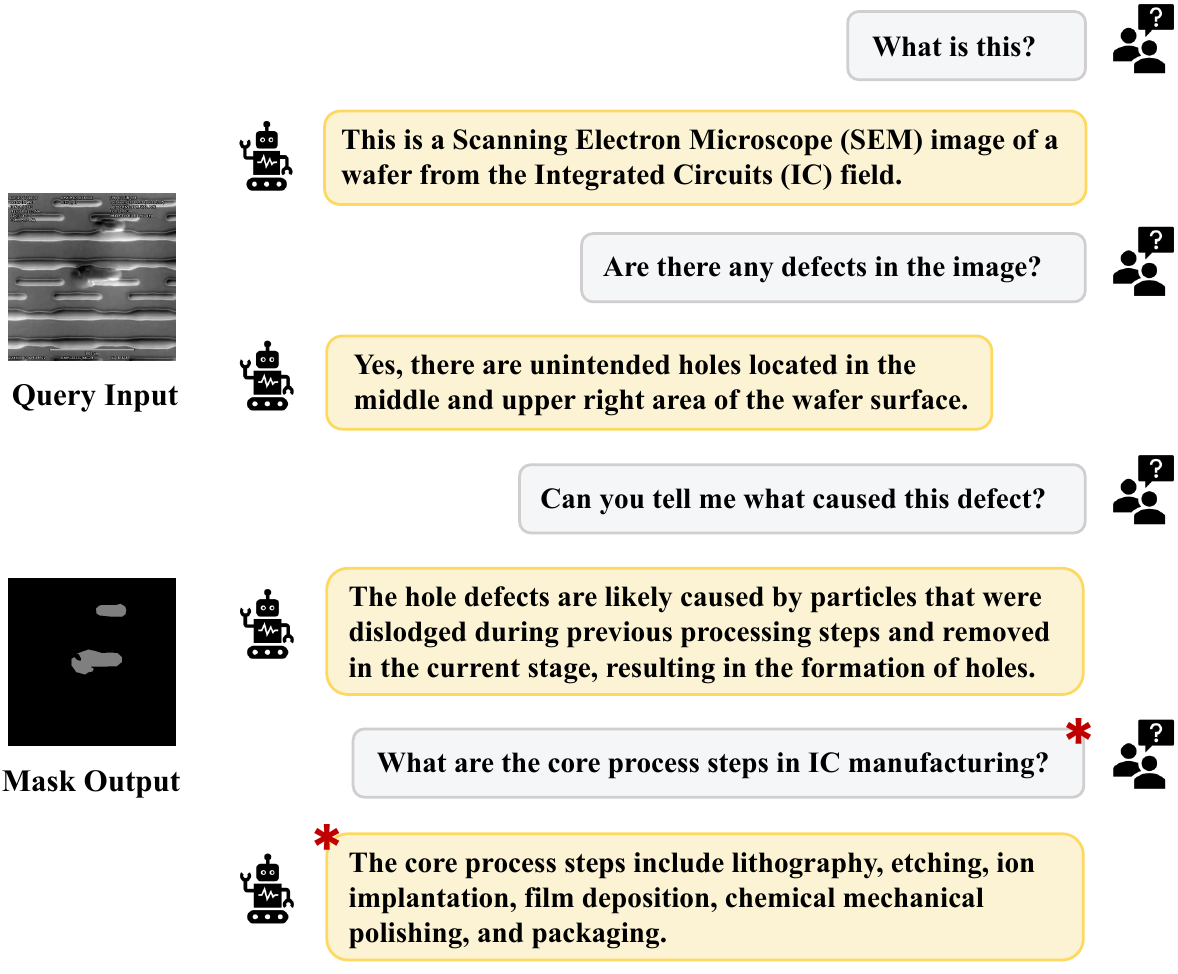}
    \caption{Dialogue example of FabGPT for ``hole'' type defects. Questions with ``\textbf{\textcolor{red2}{$\ast$}}'' are general IC questions that are not closely related to the input images.}
    \label{fig:QA}
\end{figure}

\noindent\textbf{Q\&A Task}:

According to \Cref{fig:QA}, FabGPT provides detailed defect knowledge, offering engineers valuable insights. Notably, the questions shown are not part of the training corpus. FabGPT identifies and localizes defects in SEM images, analyzing properties like "hole." It identifies root causes and suggests process solutions to optimize the production line. Additionally, FabGPT accurately answers general IC questions unrelated to the images, confirming the effectiveness of our modulation module in addressing modality bias.




\subsection{Ablation Studies}
In this section, we conduct ablation studies to demonstrate the importance of each component introduced at every stage and the operations within them.

\noindent\textbf{Stage Components}:

We study the proposed components in each stage as shown in \Cref{tab:ablations}. It confirms the importance of each individual component and the best way to structure the sub-stage. It can be seen that the prediction of the PM and pre-trained experts significantly improved defect detection in the modal enhancement stage, and the operations in the modulator played a crucial role in addressing bias issues in the Q\&A stage.

\begin{table}[t]
    \centering
    \caption{The ablation results of important components in each stage are recorded using the Pixel-AUC and the Q\&A accuracy (\%) of defect-related. The best results are \textbf{highlighted}.}
    \resizebox{\linewidth}{!}
    {
        \begin{threeparttable}
            {
                \begin{tabular}{c|c|c|c|c}
                 \toprule
                    \multirow{3}{*}{Components}   & \multirow{3}{*}{Stage} & \multicolumn{3}{c}{Task} \\
                    \cmidrule(r){3-5}
                    & & \multirow{2}{*}{Pixel-AUC}  & \multirow{2}{*}{Defect-Related}  & Alleviate \\
                    & & & & Bias \\
                    \midrule
                    + Text Mark  & Text Input & 88.61      & 83.33  & $\times$ \\
                    + PM  & Enhancement & 92.57 (+3.96\%)   & 86.67 (+3.34\%)  & $\times$ \\
                    + Pre-trained Experts& Enhancement &\textbf{95.61 (+3.04\%)}& 93.33 (+6.66\%) & $\times$ \\
                    \midrule
                    + Q-Former  & Q\&A &\textbf{95.61 (+0.00\%)}&\textbf{96.67 (+3.34\%)} & $\times$ \\
                    + Corrector  & Q\&A &\textbf{95.61 (+0.00\%)}&\textbf{96.67 (+0.00\%)} & \textbf{\checkmark}\\
                    \bottomrule
                \end{tabular}
            }
        \end{threeparttable}
    }
    \label{tab:ablations}
\end{table}

\noindent\textbf{Design for PM Operation}:

\Cref{tab:method-ablations} shows the advantages of our PM strategy. Cosine similarity outperforms matrix multiplication, and combining it with Linear and ReLU improves computation.
\begin{table}[t]
    \centering
    \caption{Ablation studies comparing different methods in the PM, with the AUC metric used to evaluate detection accuracy.}
    \resizebox{0.70\linewidth}{!}
    {
        \begin{threeparttable}
            {
                \begin{tabular}{c|c|c|c}
    \toprule
    \multicolumn{2}{c|}{Strategy}   & Image-AUC  & Pixel-AUC  \\
    \midrule
    - & \multicolumn{1}{c|}{\begin{tabular}{@{}c@{}}BiLinear \\ Similarity\end{tabular}}   & 85.77  & 89.33 \\
    \midrule
    - & \multicolumn{1}{c|}{\begin{tabular}{@{}c@{}}Matrix \\ Multiplication\end{tabular}} & 88.25  & 92.59 \\
    \midrule
    - & \multicolumn{1}{c|}{\begin{tabular}{@{}c@{}}Cosine \\ Similarity\end{tabular}}     & 91.03  & 94.61 \\
    \midrule
    \textbf{Linear + ReLU} & \multicolumn{1}{c|}{\begin{tabular}{@{}c@{}}\textbf{Cosine} \\ \textbf{Similarity}\end{tabular}}   & \textbf{91.81}  & \textbf{95.61} \\
    \bottomrule
\end{tabular}

            }
        \end{threeparttable}
    }
    \label{tab:method-ablations}
\end{table}

\noindent\textbf{Embedding Scheme of Prompt Instructions}:

\begin{table}[t]
    \centering
    \caption{Ablation experiments on the embedding schemes of visual prompt instructions to validate the Q\&A functionality.}
    \resizebox{0.78\linewidth}{!}
    {
        \begin{threeparttable}
            {
                \begin{tabular}{c|c|c}
                 \toprule
                    Visual Instruction   & Defect-Related  & Unrelated  \\
                    \midrule
                    $T_{img}$ & 82.50   & 10.00 \\
                    $T_{img} + T_{txt}$ & 85.00   & 10.00 \\
                    $T_{img} + T_{txt} + T_{mas}$ & 95.00   & 10.00 \\
                    $T_{vis} + T_{mas}$ & \textbf{96.67}   & 12.00 \\
                    $\textcolor{red}{a}T_{vis} + T_{mas}$ & \textbf{96.67}   & \textbf{98.00} \\
                    \bottomrule
                \end{tabular}
            }
        \end{threeparttable}
    }
    \label{tab:prompt-ablations}
\end{table}

\Cref{tab:prompt-ablations} shows that our prompt embedding scheme effectively mitigates modality bias. By updating the corresponding coefficient ``$a$'' for visual tokens, our model effectively discerns the relationship between user queries and visual inputs and mitigates the modality bias issues.

\section{Conclusions}
\label{sec:conclusions}

In this paper, we introduce a novel large multimodal language model, FabGPT, for defect knowledge querying in the IC field, including defection, analysis, Q\&A, \textit{etc}. It employs three stages to gradually achieve the functionality of defect detection and high-quality dialogue. 
We validate the effectiveness of FabGPT on the SEM-WaD dataset and 100 questions. Our work provides great convenience for the semiconductor industry and also offers insights for further LMM research.

\section*{Acknowledgments}
This work is sponsored by The National Natural Science Foundation of China (Grant No.~62034007).

\clearpage
{

}


\begin{thebibliography}{10}
\providecommand{\url}[1]{#1}
\csname url@samestyle\endcsname
\providecommand{\newblock}{\relax}
\providecommand{\bibinfo}[2]{#2}
\providecommand{\BIBentrySTDinterwordspacing}{\spaceskip=0pt\relax}
\providecommand{\BIBentryALTinterwordstretchfactor}{4}
\providecommand{\BIBentryALTinterwordspacing}{\spaceskip=\fontdimen2\font plus
\BIBentryALTinterwordstretchfactor\fontdimen3\font minus \fontdimen4\font\relax}
\providecommand{\BIBforeignlanguage}[2]{{%
\expandafter\ifx\csname l@#1\endcsname\relax
\typeout{** WARNING: IEEEtran.bst: No hyphenation pattern has been}%
\typeout{** loaded for the language `#1'. Using the pattern for}%
\typeout{** the default language instead.}%
\else
\language=\csname l@#1\endcsname
\fi
#2}}
\providecommand{\BIBdecl}{\relax}
\BIBdecl

\bibitem{achiam2023gpt}
J.~Achiam, S.~Adler, S.~Agarwal, L.~Ahmad, I.~Akkaya, F.~L. Aleman, D.~Almeida, J.~Altenschmidt, S.~Altman, S.~Anadkat \emph{et~al.}, ``Gpt-4 technical report,'' \emph{arXiv preprint arXiv:2303.08774}, 2023.

\bibitem{su2023pandagpt}
Y.~Su, T.~Lan, H.~Li, J.~Xu, Y.~Wang, and D.~Cai, ``Pandagpt: One model to instruction-follow them all,'' \emph{arXiv preprint arXiv:2305.16355}, 2023.

\bibitem{touvron2023llama}
H.~Touvron, T.~Lavril, G.~Izacard, X.~Martinet, M.-A. Lachaux, T.~Lacroix, B.~Rozi{\`e}re, N.~Goyal, E.~Hambro, F.~Azhar \emph{et~al.}, ``Llama: Open and efficient foundation language models,'' \emph{arXiv preprint arXiv:2302.13971}, 2023.

\bibitem{zhu2023minigpt}
D.~Zhu, J.~Chen, X.~Shen, X.~Li, and M.~Elhoseiny, ``Minigpt-4: Enhancing vision-language understanding with advanced large language models,'' \emph{arXiv preprint arXiv:2304.10592}, 2023.

\bibitem{radford2021learning}
A.~Radford, J.~W. Kim, C.~Hallacy, A.~Ramesh, G.~Goh, S.~Agarwal, G.~Sastry, A.~Askell, P.~Mishkin, J.~Clark \emph{et~al.}, ``Learning transferable visual models from natural language supervision,'' in \emph{International conference on machine learning}.\hskip 1em plus 0.5em minus 0.4em\relax PMLR, 2021, pp. 8748--8763.

\bibitem{reddy2021dall}
M.~D.~M. Reddy, M.~S.~M. Basha, M.~M.~C. Hari, and M.~N. Penchalaiah, ``Dall-e: Creating images from text,'' \emph{UGC Care Group I Journal}, vol.~8, no.~14, pp. 71--75, 2021.

\bibitem{gu2023pre}
Y.~Gu, L.~Dong, F.~Wei, and M.~Huang, ``Pre-training to learn in context,'' \emph{arXiv preprint arXiv:2305.09137}, 2023.

\bibitem{chen2022improving}
M.~Chen, J.~Du, R.~Pasunuru, T.~Mihaylov, S.~Iyer, V.~Stoyanov, and Z.~Kozareva, ``Improving in-context few-shot learning via self-supervised training,'' \emph{arXiv preprint arXiv:2205.01703}, 2022.

\bibitem{wei2023symbol}
J.~Wei, L.~Hou, A.~Lampinen, X.~Chen, D.~Huang, Y.~Tay, X.~Chen, Y.~Lu, D.~Zhou, T.~Ma \emph{et~al.}, ``Symbol tuning improves in-context learning in language models,'' \emph{arXiv preprint arXiv:2305.08298}, 2023.

\bibitem{quirk2001semiconductor}
M.~Quirk and J.~Serda, \emph{Semiconductor manufacturing technology}.\hskip 1em plus 0.5em minus 0.4em\relax Prentice Hall Upper Saddle River, NJ, 2001, vol.~1.

\bibitem{fan2019key}
S.-K.~S. Fan, D.-M. Tsai, F.~He, J.-Y. Huang, and C.-H. Jen, ``Key parameter identification and defective wafer detection of semiconductor manufacturing processes using image processing techniques,'' \emph{IEEE Transactions on Semiconductor Manufacturing}, vol.~32, no.~4, pp. 544--552, 2019.

\bibitem{lechien2023automated}
T.~Lechien, E.~Dehaerne, B.~Dey, V.~Blanco, S.~De~Gendt, and W.~Meert, ``Automated semiconductor defect inspection in scanning electron microscope images: a systematic review,'' \emph{arXiv preprint arXiv:2308.08376}, 2023.

\bibitem{seebauer2010trends}
E.~G. Seebauer and K.~W. Noh, ``Trends in semiconductor defect engineering at the nanoscale,'' \emph{Materials Science and Engineering: R: Reports}, vol.~70, no. 3-6, pp. 151--168, 2010.

\bibitem{gu2023anomalygpt}
Z.~Gu, B.~Zhu, G.~Zhu, Y.~Chen, M.~Tang, and J.~Wang, ``Anomalygpt: Detecting industrial anomalies using large vision-language models,'' \emph{arXiv preprint arXiv:2308.15366}, 2023.

\bibitem{pang2021explainable}
G.~Pang, C.~Ding, C.~Shen, and A.~v.~d. Hengel, ``Explainable deep few-shot anomaly detection with deviation networks,'' \emph{arXiv preprint arXiv:2108.00462}, 2021.

\bibitem{cheon2019convolutional}
S.~Cheon, H.~Lee, C.~O. Kim, and S.~H. Lee, ``Convolutional neural network for wafer surface defect classification and the detection of unknown defect class,'' \emph{IEEE Transactions on Semiconductor Manufacturing}, vol.~32, no.~2, pp. 163--170, 2019.

\bibitem{ding2022catching}
C.~Ding, G.~Pang, and C.~Shen, ``Catching both gray and black swans: Open-set supervised anomaly detection,'' in \emph{Proceedings of the IEEE/CVF conference on computer vision and pattern recognition}, 2022, pp. 7388--7398.

\bibitem{zhang2023prototypical}
H.~Zhang, Z.~Wu, Z.~Wang, Z.~Chen, and Y.-G. Jiang, ``Prototypical residual networks for anomaly detection and localization,'' in \emph{Proceedings of the IEEE/CVF Conference on Computer Vision and Pattern Recognition}, 2023, pp. 16\,281--16\,291.

\bibitem{yao2023explicit}
X.~Yao, R.~Li, J.~Zhang, J.~Sun, and C.~Zhang, ``Explicit boundary guided semi-push-pull contrastive learning for supervised anomaly detection,'' in \emph{Proceedings of the IEEE/CVF Conference on Computer Vision and Pattern Recognition}, 2023, pp. 24\,490--24\,499.

\bibitem{lai2023lisa}
X.~Lai, Z.~Tian, Y.~Chen, Y.~Li, Y.~Yuan, S.~Liu, and J.~Jia, ``Lisa: Reasoning segmentation via large language model,'' \emph{arXiv preprint arXiv:2308.00692}, 2023.

\bibitem{zontak2009kernel}
M.~Zontak and I.~Cohen, ``Kernel-based detection of defects on semiconductor wafers,'' in \emph{2009 IEEE international workshop on machine learning for signal processing}.\hskip 1em plus 0.5em minus 0.4em\relax IEEE, 2009, pp. 1--6.

\bibitem{gomez2022optimal}
J.~L. G{\'o}mez-Sirvent, F.~L. de~la Rosa, R.~S{\'a}nchez-Reolid, A.~Fern{\'a}ndez-Caballero, and R.~Morales, ``Optimal feature selection for defect classification in semiconductor wafers,'' \emph{IEEE Transactions on Semiconductor Manufacturing}, vol.~35, no.~2, pp. 324--331, 2022.

\bibitem{li2023blip}
J.~Li, D.~Li, S.~Savarese, and S.~Hoi, ``Blip-2: Bootstrapping language-image pre-training with frozen image encoders and large language models,'' in \emph{International conference on machine learning}.\hskip 1em plus 0.5em minus 0.4em\relax PMLR, 2023, pp. 19\,730--19\,742.

\bibitem{chiang2023vicuna}
W.-L. Chiang, Z.~Li, Z.~Lin, Y.~Sheng, Z.~Wu, H.~Zhang, L.~Zheng, S.~Zhuang, Y.~Zhuang, J.~E. Gonzalez \emph{et~al.}, ``Vicuna: An open-source chatbot impressing gpt-4 with 90\%* chatgpt quality,'' \emph{See https://vicuna. lmsys. org (accessed 14 April 2023)}, vol.~2, no.~3, p.~6, 2023.

\bibitem{alaparthi2020bidirectional}
S.~Alaparthi and M.~Mishra, ``Bidirectional encoder representations from transformers (bert): A sentiment analysis odyssey,'' \emph{arXiv preprint arXiv:2007.01127}, 2020.

\bibitem{yuan2021tokens}
L.~Yuan, Y.~Chen, T.~Wang, W.~Yu, Y.~Shi, Z.-H. Jiang, F.~E. Tay, J.~Feng, and S.~Yan, ``Tokens-to-token vit: Training vision transformers from scratch on imagenet,'' in \emph{Proceedings of the IEEE/CVF international conference on computer vision}, 2021, pp. 558--567.

\bibitem{rebuffi2017learning}
S.-A. Rebuffi, H.~Bilen, and A.~Vedaldi, ``Learning multiple visual domains with residual adapters,'' \emph{Advances in neural information processing systems}, vol.~30, 2017.

\bibitem{li2022pp}
C.~Li, W.~Liu, R.~Guo, X.~Yin, K.~Jiang, Y.~Du, Y.~Du, L.~Zhu, B.~Lai, X.~Hu \emph{et~al.}, ``Pp-ocrv3: More attempts for the improvement of ultra lightweight ocr system,'' \emph{arXiv preprint arXiv:2206.03001}, 2022.

\bibitem{girdhar2023imagebind}
R.~Girdhar, A.~El-Nouby, Z.~Liu, M.~Singh, K.~V. Alwala, A.~Joulin, and I.~Misra, ``Imagebind: One embedding space to bind them all,'' in \emph{Proceedings of the IEEE/CVF Conference on Computer Vision and Pattern Recognition}, 2023, pp. 15\,180--15\,190.

\bibitem{huang2019ccnet}
Z.~Huang, X.~Wang, L.~Huang, C.~Huang, Y.~Wei, and W.~Liu, ``Ccnet: Criss-cross attention for semantic segmentation,'' in \emph{Proceedings of the IEEE/CVF international conference on computer vision}, 2019, pp. 603--612.

\bibitem{lin2017focal}
T.-Y. Lin, P.~Goyal, R.~Girshick, K.~He, and P.~Doll{\'a}r, ``Focal loss for dense object detection,'' in \emph{Proceedings of the IEEE international conference on computer vision}, 2017, pp. 2980--2988.

\bibitem{milletari2016v}
F.~Milletari, N.~Navab, and S.-A. Ahmadi, ``V-net: Fully convolutional neural networks for volumetric medical image segmentation,'' in \emph{2016 fourth international conference on 3D vision (3DV)}.\hskip 1em plus 0.5em minus 0.4em\relax Ieee, 2016, pp. 565--571.

\bibitem{loshchilov2017decoupled}
I.~Loshchilov and F.~Hutter, ``Decoupled weight decay regularization,'' \emph{arXiv preprint arXiv:1711.05101}, 2017.

\bibitem{loshchilov2016sgdr}
------, ``Sgdr: Stochastic gradient descent with warm restarts,'' \emph{arXiv preprint arXiv:1608.03983}, 2016.

\end{thebibliography}
\end{document}